\documentclass[pdflatex,sn-nature]{sn-jnl}

\usepackage{graphicx}%
\usepackage{multirow}%
\usepackage{multicol}
\usepackage{amsmath,amssymb,amsfonts}%
\usepackage{booktabs}%
\usepackage{array}%
\usepackage{longtable}%
\usepackage{xcolor}%
\usepackage{caption}%
\usepackage{subcaption}%
\usepackage{adjustbox}%

\begin{document}

\title[Graph Benchmarking for Synthesis Planning]{ProvMind: Provenance-grounded reasoning for materials synthesis}

\author[1]{\fnm{Yiming} \sur{Zhang}}
\author[1,2]{\fnm{Ryo} \sur{Tamura}}\email{tamura.ryo@nims.go.jp}
\author*[1,2,3]{\fnm{Koji} \sur{Tsuda}}\email{tsuda@k.u-tokyo.ac.jp}

\affil[1]{\orgdiv{Graduate School of Frontier Sciences}, \orgname{The University of Tokyo}, \orgaddress{\city{Chiba}, \country{Japan}}}
\affil[2]{\orgname{Center for Basic Research on Materials, National Institute for Materials Science}, \orgaddress{\city{Tsukuba}, \state{Ibaraki}, \country{Japan}}}
\affil[3]{\orgname{RIKEN Center for Advanced Intelligence Project}, \orgaddress{\city{Tokyo}, \country{Japan}}}

\abstract {
Materials process optimization requires reasoning over routes, conditions, tools and causal dependencies, yet most computational formulations flatten synthesis procedures into text or ordered steps. We introduce MatProcBench, a provenance-grounded benchmark constructed from literature-mined MatPROV graphs, to evaluate seven process-reasoning tasks spanning route continuity, step-level variable inference and global causal consistency under both same-split and shift-aware evaluation, including a strict dual-OOD split that combines temporal and material-class shift. We further introduce ProvMind, a process-memory reasoning framework that retrieves analogous training processes, converts them into provenance-aware option-level compatibility scores, and uses a language model for constrained final decision making. ProvMind achieves 52.84\% accuracy on the dual-OOD split, outperforming prompting, retrieval-augmented and supervised fine-tuning baselines.
}

\maketitle
\section{Introduction}
Computational materials research is increasingly shifting from the prediction of static material properties toward the control of experimental processes that determine whether a candidate material can be realized, refined and deployed \citep{wang2023scientific, boiko2023autonomous, stach2021autonomous, szymanski2023autonomous, abolhasani2023rise, cheetham2022chemical}. In this setting, the central scientific question is no longer limited to whether a target composition is synthesizable in principle, but extends to how a process should be interpreted, compared and modified. Experimentalists routinely need to determine which operation governs a downstream transformation, which precursor or intermediate propagates through a synthesis route, which tool constrains feasibility, and which processing variable explains divergence among related experiments. These are fundamentally questions of materials process informatics. Addressing them requires reasoning over procedures as structured experimental systems rather than as loosely ordered textual descriptions \citep{kim2017materials, kononova2019text, wang2022dataset, kovnir2021predictive, muraoka2019linking}.

This distinction is consequential because optimization-relevant information in materials synthesis is inherently relational. A synthesis process is not merely a sequence of operations, but a directed flow of matter through activities, intermediate states, tools and environmental conditions linked by causal dependencies. These dependencies determine which variables may be altered, which constraints must be preserved and which interventions remain physically meaningful. A milling operation may modify the state of a precursor consumed by a later sintering step; a processing atmosphere may constitute the sole experimental variable distinguishing two otherwise similar routes; a tool may participate only within a specific branch of a synthesis workflow. Linear recipe representations frequently obscure these relations and weaken explicit causal traceability \citep{kononova2019text, wang2022dataset, muraoka2019linking}, while recent condition-centric synthesis studies further underscore the importance of preserving process-level dependencies across steps \citep{huo2022machine, karpovich2023interpretable, wang2024optimal}. For process optimization, this loss is not cosmetic. It removes the structural information required to localize interventions and reason about how their effects propagate through a synthesis route.

Graph structure therefore plays a foundational role in process informatics, although in the present study it is treated primarily as a representational substrate for scientific reasoning rather than as an end in itself. MatPROV is particularly valuable in this regard because it represents synthesis procedures as provenance graphs with typed material, activity and tool nodes linked through explicit Usage and Generation relations \citep{tsuruta2025matprov}. This organization distinguishes fresh precursors from propagated intermediates, consumed tools from generated materials, and local adjacency from true causal precedence. Several reasoning operations required for process informatics, including route reconstruction and causally valid process ordering, depend directly on this topology, whereas others, including condition inference and tool identification, rely on typed associations between activities and their material or equipment context. Provenance structure therefore renders process reasoning operationally well defined, in a manner consistent with long-standing calls for more explicit synthesis knowledge representations and route-centered planning in materials science \citep{jansen2002concept, aykol2021rational}.

Existing computational formulations only partially address this requirement. Prior work in materials synthesis has advanced precursor recommendation, synthesis prediction, route ranking and language-model-based procedural generation \citep{kim2020inorganic, aykol2021rational, huo2022machine, he2023precursor, karpovich2023interpretable, kim2024predicting, kim2024large, noh2024retrieval, prein2025language, prein2025retro, noh2026msp}. Nevertheless, many existing approaches continue to reduce synthesis procedures to flat text, isolated attributes or ordered step sequences. Even retrieval-augmented formulations typically provide static contextual evidence rather than explicit mechanisms for provenance-grounded reasoning \citep{lewis2020retrieval, m2024augmenting, jin2024graph, hu2025grag, peng2025graph}. As a result, it remains difficult to determine whether a model genuinely captures the causal and variable-bearing structure of a materials process or instead relies primarily on local lexical similarity. What remains missing is both a benchmark formulation that isolates the core reasoning operations underlying process interpretation and an inference framework capable of exploiting provenance structure under realistic distribution shift, a gap that echoes broader concerns about the promises and current limitations of AI for synthesis research \citep{david2023promise}.

We address this gap by formulating provenance-grounded materials process reasoning as a benchmarkable scientific problem. We introduce MatProcBench, a benchmark derived from MatPROV provenance records and designed to evaluate whether models can recover route continuity, process variables and causal consistency from experimentally structured evidence. The benchmark includes route retrieval, missing-step identification, next-activity prediction, condition prediction, full-condition-set prediction, tool selection and process ordering, spanning both local process understanding and global causal reasoning. To evaluate robustness under realistic scientific shift, MatProcBench incorporates both same-split and cross-split evaluation protocols, including a strict dual-distribution-shift setting that simultaneously enforces temporal and material-class separation. Supplementary Note~1 summarizes the benchmark composition, split statistics and distribution-shift structure in greater detail.

\begin{figure}[t]
\centering
\includegraphics[width=\textwidth]{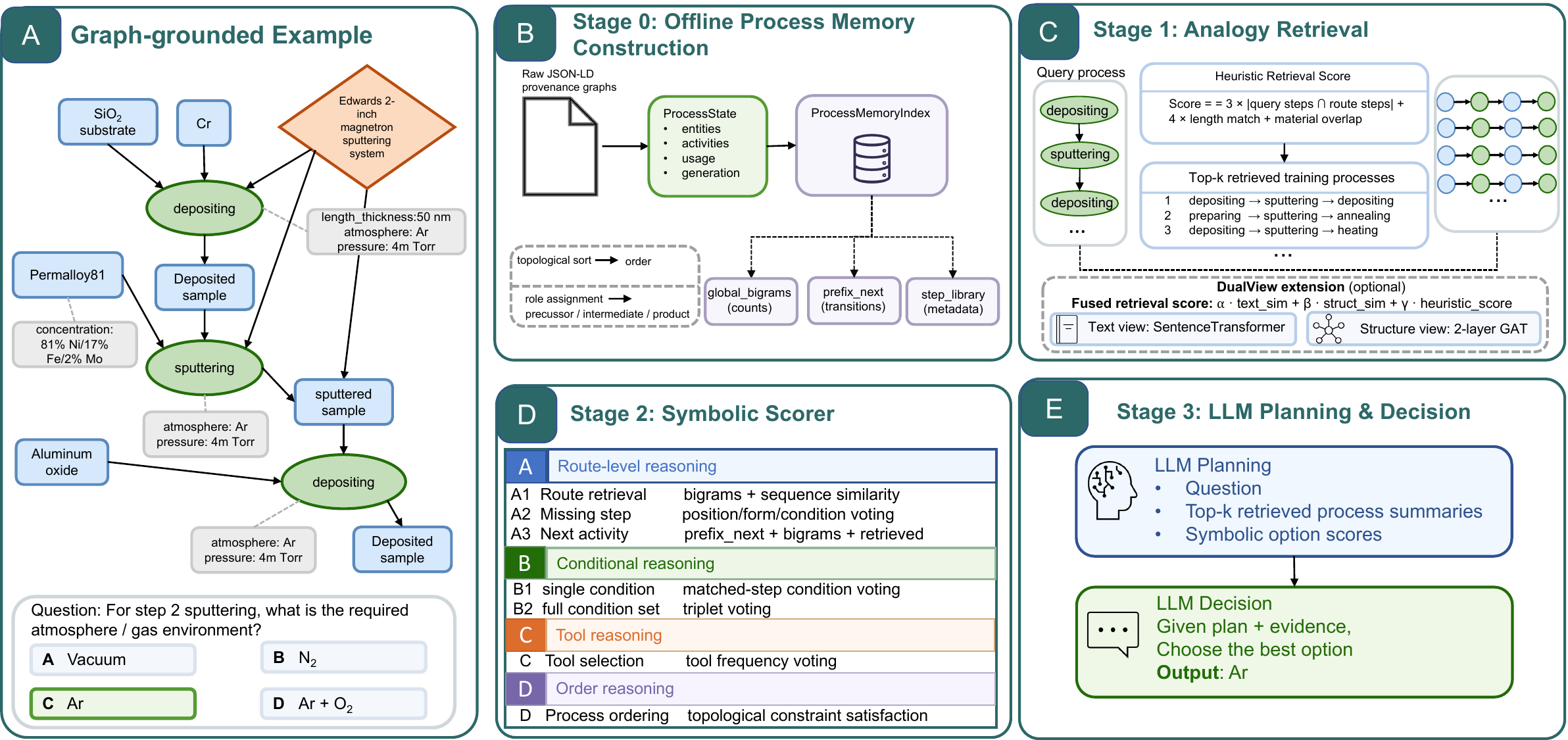}
\caption{Overview of the ProvMind framework and the provenance-grounded reasoning setting addressed in this study. \textbf{A}, Graph-grounded example benchmark instance illustrating that the correct answer depends on typed process context and causal provenance rather than on local lexical cues alone. \textbf{B}, Offline process-memory construction from training-split MatPROV graphs, including topological ordering, semantic role assignment and reusable route- and step-level statistics. \textbf{C}, Analogy retrieval at inference time, where symbolic, text-based and structure-aware views are fused to identify experimentally compatible precedents. \textbf{D}, Task-aware symbolic scoring that converts retrieved precedents into option-level compatibility evidence for route, condition, tool and ordering tasks. \textbf{E}, Final language-model planning and decision making conditioned on the query, retrieved process summaries and provenance-grounded scores.}
\label{fig:overview}
\end{figure}

To study this problem setting, we further introduce ProvMind, a provenance-grounded reasoning framework that retrieves analogous training processes from process memory, converts them into structured compatibility evidence, and performs inference using provenance-aware symbolic support \citep{aamodt1994case, zeyen2018conversational}. Rather than relying solely on parametric recall or static retrieval augmentation, ProvMind explicitly compares candidate processes against structurally related experimental precedents derived from the training split. This formulation reflects the hypothesis that robust reasoning about materials processes under scientific distribution shift depends primarily on provenance-grounded structural comparison rather than neural similarity alone. Figure~\ref{fig:overview} summarizes the full inference pipeline, from provenance-graph compilation and process-memory construction to analogy retrieval, task-aware symbolic scoring and constrained language-model decision making.

Across both matched and shift-aware evaluations, provenance-grounded process memory consistently outperforms prompting, static retrieval augmentation and supervised fine-tuning baselines. The strongest gains emerge under realistic distribution shift, where provenance-aware symbolic structure provides substantially stronger generalization than neural similarity alone. We further observe that random partitioning markedly inflates performance on combinatorial process reasoning tasks, whereas shift-aware evaluation exposes persistent bottlenecks in global route discrimination and joint condition recovery. Together, these findings position materials process informatics as a distinct computational reasoning problem grounded in experimentally structured provenance graphs and demonstrate the importance of process-memory reasoning for robust scientific inference.

\section{Results}
\subsection{Same-Split Performance Under Distribution Shift}

\begin{figure}[t]
\centering
\includegraphics[width=\textwidth]{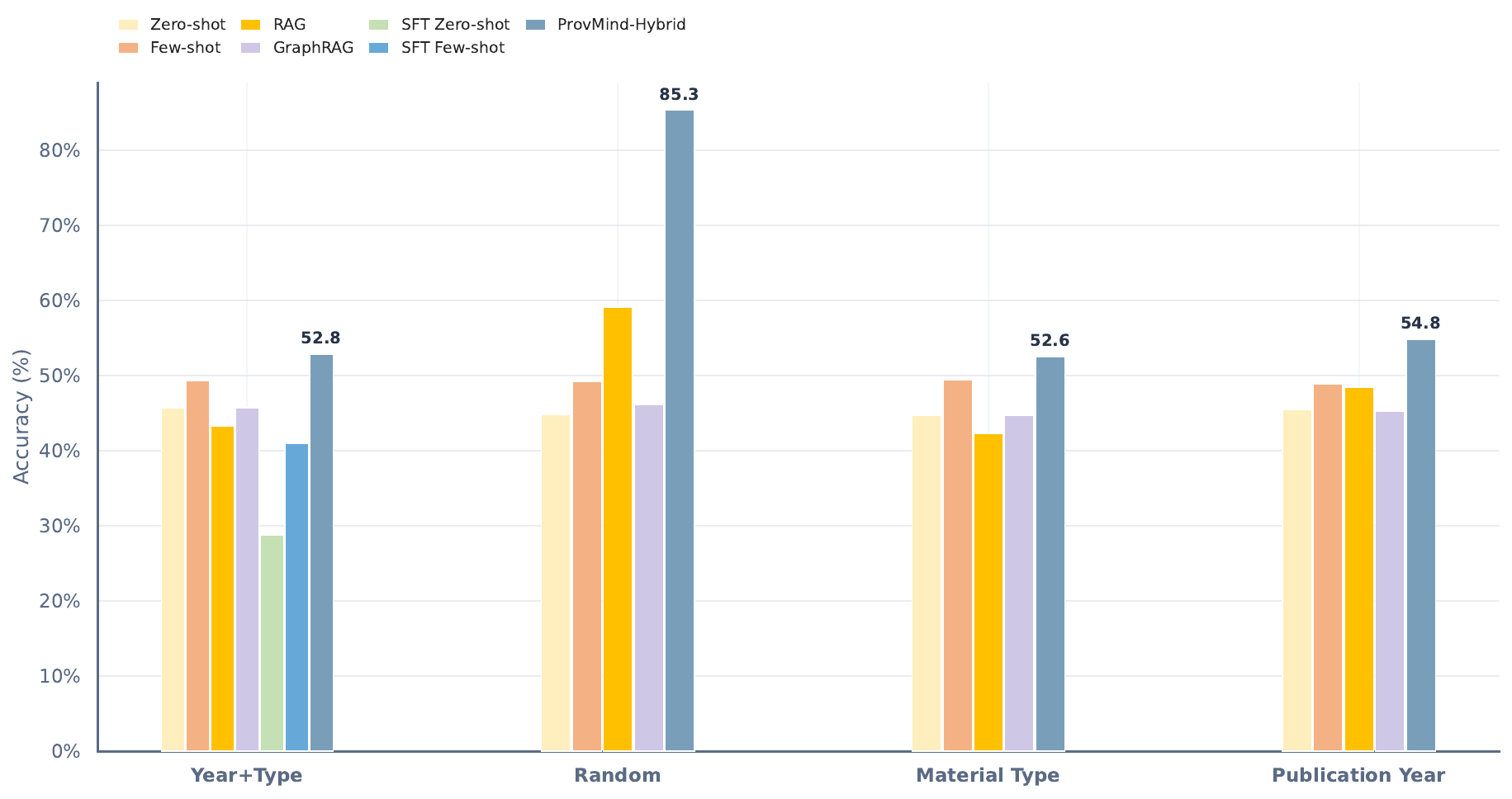}
\caption{Overall same-split benchmark performance across evaluation protocols. Bars report accuracy for prompting, retrieval-augmented, graph-aware retrieval, supervised fine-tuning and ProvMind variants under the dual-distribution-shift split (Year+Type), random split, material-type split and publication-year split. ProvMind-Hybrid is the strongest method on all shift-aware settings, whereas the random split yields substantially inflated accuracies for several approaches and is therefore markedly less diagnostic of robust process reasoning.}
\label{fig:same_split_overview}
\end{figure}

We first evaluated prompting methods, retrieval-based baselines and ProvMind under same-split evaluation using Qwen2.5-7B-Instruct. As summarized in Fig.~\ref{fig:same_split_overview}, provenance-grounded process memory is the only approach that remains consistently strong across all evaluation regimes while avoiding the instability shown by purely neural retrieval methods. Under the dual-distribution-shift setting, ProvMind achieved 52.84\% accuracy, exceeding the strongest non-agentic baseline, few-shot prompting, by 3.51 percentage points (49.33\%). Static retrieval approaches were substantially less effective in this regime, with RAG and GraphRAG reaching 43.32\% and 45.70\%, respectively. These observations indicate that attaching retrieved precedents or local process neighbourhoods is insufficient when both temporal and material-class overlap are explicitly constrained. Full same-split and cross-split result tables are provided in Supplementary Note~2.

The same pattern persisted across the two single-axis shift settings. On the material-type split, ProvMind achieved 52.55\%, outperforming few-shot prompting by 3.13 points (49.42\%). On the publication-year split, the margin increased further, with ProvMind reaching 54.82\%, compared with 48.95\% for few-shot prompting and 48.50\% for RAG. Across all shift-aware evaluations, provenance-grounded process memory consistently produced the strongest performance.

By contrast, random partitioning yielded substantially inflated accuracies for multiple methods. ProvMind reached 85.33\% under the random split, while RAG increased to 59.12\%. The large divergence between the random partition and the shift-aware settings suggests that random splitting substantially relaxes the benchmark and overestimates practical generalization performance for process reasoning tasks.

\subsection{Cross-Split Generalization from Dual-Distribution-Shift Process Memory}

\begin{figure}[t]
\centering
\includegraphics[width=\textwidth]{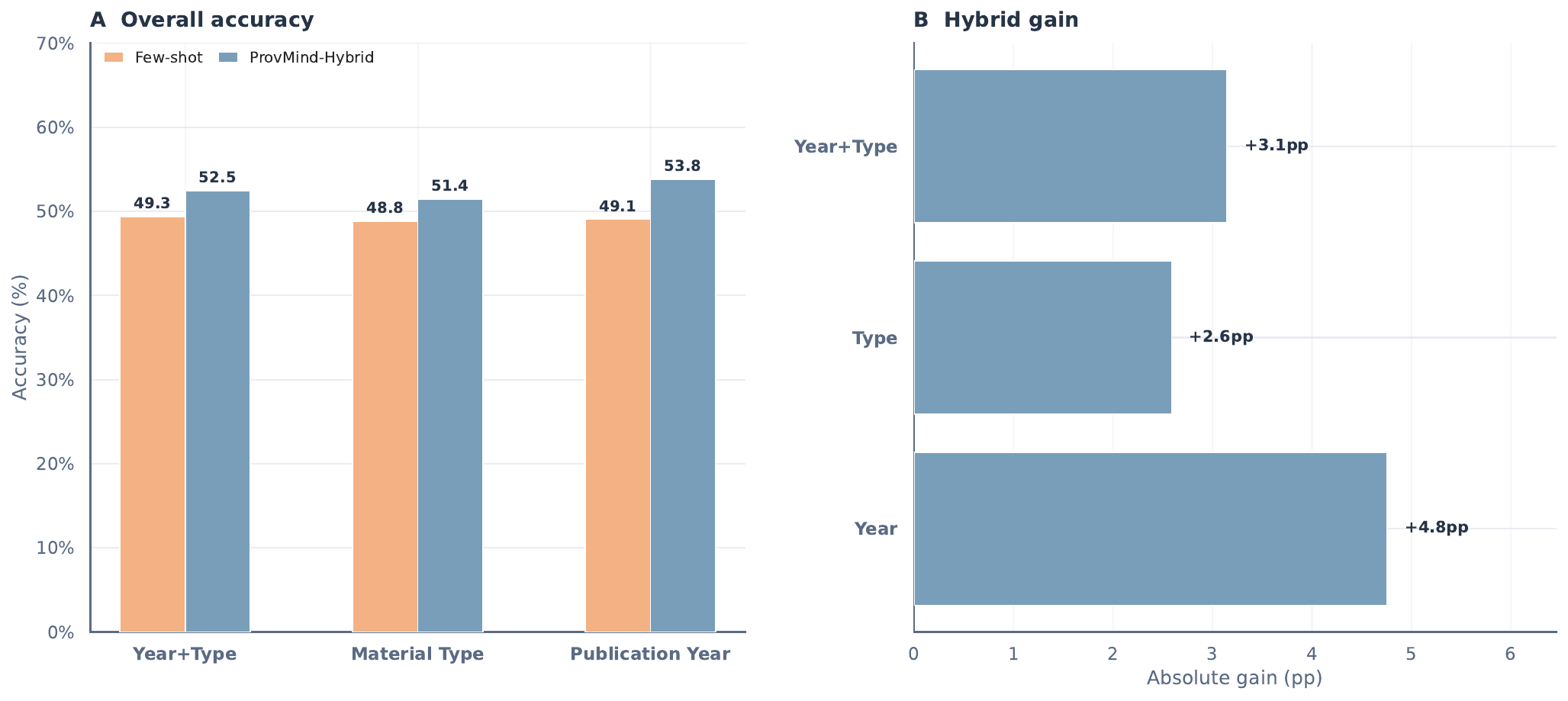}
\caption{Cross-split generalization from the dual-distribution-shift training regime. \textbf{A}, Overall accuracy when train-dependent components are fixed on the strictest split and evaluated on the dual-distribution-shift (Year+Type), material-type and publication-year test partitions; ProvMind-Hybrid is compared against the strongest learned reference, SFT zero-shot. \textbf{B}, Absolute gain of ProvMind-Hybrid over SFT zero-shot in each clean evaluation setting, showing that provenance-grounded process memory preserves a positive margin under all non-random distribution shifts.}
\label{fig:ood_generalization}
\end{figure}

We next fixed all train-dependent components using the dual-distribution-shift training split and evaluated transfer to alternative test partitions. This protocol isolates whether a framework trained or indexed under the strictest distributional regime can generalize across unseen temporal and material-class conditions. Figure~\ref{fig:ood_generalization} summarizes both the absolute accuracies and the residual advantage retained by ProvMind-Hybrid over the strongest learned baseline after this train--test decoupling.

Among prompting-based baselines, few-shot prompting remained the strongest conventional reference, achieving 48.84\% on the material-type split and 49.05\% on the publication-year split. Static retrieval again failed to preserve robustness: RAG decreased to 42.27\% and 43.86\% on the material-type and publication-year splits, respectively, while GraphRAG remained near the zero-shot baseline at 44.94\% and 45.35\%.

The strongest learned baseline in this setting was the SFT zero-shot model trained on the dual-distribution-shift split. It achieved 50.10\% on the matched dual-distribution-shift test set, 49.60\% on the material-type split and 50.21\% on the publication-year split. ProvMind-Hybrid exceeded this learned reference across all clean evaluation settings, reaching 52.48\%, 51.43\% and 53.81\%, respectively. The corresponding gains over SFT zero-shot were +2.38, +1.83 and +3.60 percentage points.

The persistence of this advantage across all clean evaluation settings indicates that provenance-grounded process memory transfers beyond matched train--test distributions. These results further suggest that robust process reasoning under scientific shift depends primarily on structurally grounded experimental comparison rather than parameter adaptation alone.

\subsection{Task-Level Performance and Reasoning Difficulty}

\begin{figure}[t]
\centering
\includegraphics[width=\textwidth]{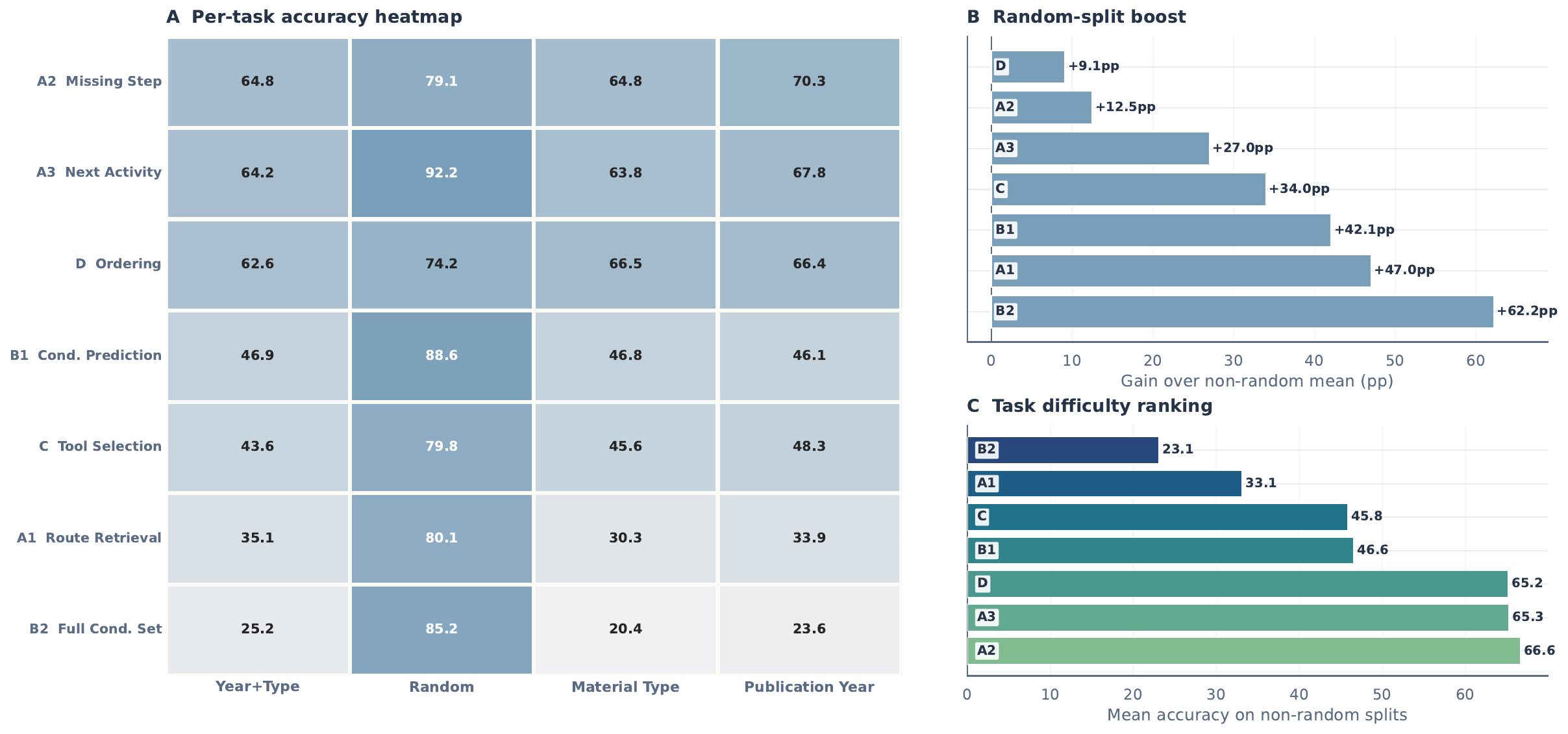}
\caption{Task hierarchy and split sensitivity. \textbf{A}, Per-task accuracy heatmap for ProvMind-Hybrid across dual-distribution-shift (Year+Type), random, material-type and publication-year evaluation, together with the mean over the three non-random splits. \textbf{B}, Accuracy boost induced by random splitting relative to the non-random mean, showing that route retrieval (A1) and full-condition-set prediction (B2) are disproportionately sensitive to hidden overlap. \textbf{C}, Task ranking by mean accuracy on the non-random splits, separating globally constrained reasoning tasks from more locally grounded ones.}
\label{fig:task_hierarchy_shift}
\end{figure}

Task-level accuracies revealed a stable hierarchy across the three non-random same-split settings. Figure~\ref{fig:task_hierarchy_shift}A shows that this hierarchy is not a single-split artifact: the same broad ranking recurs across dual-distribution-shift, material-type and publication-year evaluation, despite substantial changes in absolute difficulty. Averaged over the dual-distribution-shift, material-type and publication-year evaluations, the highest-performing tasks were A2 missing-step identification (66.62\%), A3 next-activity prediction (65.25\%) and D process ordering (65.15\%). These tasks depend primarily on local route continuity or explicit causal ordering, both of which are strongly represented within provenance-grounded process memory.

Intermediate difficulty was observed for B1 condition prediction (46.57\%) and C tool selection (45.83\%). Both tasks require step-local attribute inference while remaining anchored to a specific target step. The principal bottlenecks were A1 route retrieval and B2 full condition-set prediction, with non-random average accuracies of 33.11\% and 23.06\%, respectively. A1 requires discrimination among globally plausible multi-step routes, whereas B2 requires recovery of a jointly consistent temperature--duration--atmosphere configuration, resulting in substantially greater combinatorial complexity.

This hierarchy persisted under cross-split evaluation from dual-distribution-shift training. For ProvMind-Hybrid, task D remained robust across all clean test sets, reaching 61.07\%, 66.51\% and 66.67\% on the dual-distribution-shift, material-type and publication-year splits, respectively. Similarly, A2 achieved 62.14\%, 62.28\% and 67.34\%. In contrast, A1 remained comparatively weak at 29.73\%, 27.39\% and 27.17\%, while B2 remained the most difficult task overall at 27.18\%, 21.56\% and 24.39\%.

The persistence of this ordering under distribution shift suggests that the principal unresolved challenge lies in global route discrimination and joint condition recovery rather than local provenance grounding.

The random split again exhibited markedly different behaviour. Relative to the non-random average, random-split accuracy increased by 47.00 points on A1 and by 62.19 points on B2. As highlighted in Fig.~\ref{fig:task_hierarchy_shift}B, these two tasks are especially sensitive to hidden overlap in route patterns and condition templates, whereas Fig.~\ref{fig:task_hierarchy_shift}C makes clear that the hardest tasks under non-random evaluation are precisely those that require global discrimination or joint consistency rather than local step matching. This divergence further supports the importance of shift-aware evaluation.

\subsection{Component Ablations and Retrieval Variants}

\begin{figure}[t]
\centering
\includegraphics[width=\textwidth]{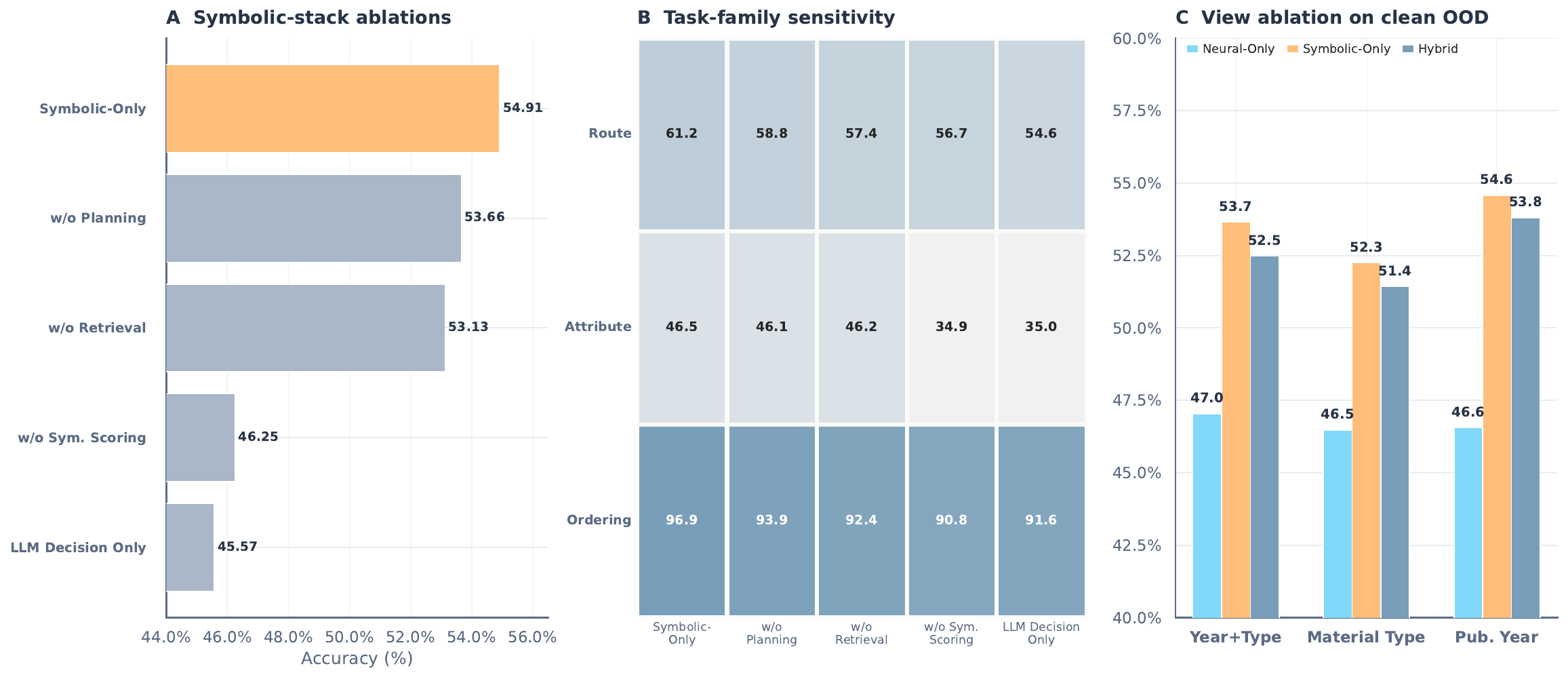}
\caption{Mechanistic interpretation of ProvMind. \textbf{A}, Symbolic-stack ablations under the dual-distribution-shift setting show that provenance-aware symbolic scoring is the dominant contributor to overall accuracy. \textbf{B}, Task-family sensitivity across ablation settings indicates that attribute inference is most vulnerable to removing symbolic structure, whereas causal ordering remains comparatively robust. \textbf{C}, Retrieval-view ablation on the clean OOD settings (Year+Type, material type and publication year) compares neural-only, symbolic-only and hybrid views; symbolic retrieval is the strongest single view, while the hybrid retains a stable advantage over neural-only retrieval across all shifts.}
\label{fig:mechanistic_ablations}
\end{figure}

We first ablated the symbolic-stack formulation under the dual-distribution-shift setting to identify the primary contributors to performance. Figure~\ref{fig:mechanistic_ablations}A provides the clearest mechanistic summary: the dominant source of robustness is not planning overhead or retrieval in the abstract, but the structured compatibility computation introduced by provenance-aware symbolic scoring. The full symbolic-stack system achieved 54.91\% accuracy. Removing planning reduced performance modestly to 53.66\% (-1.25 points), while removing retrieval produced a similar decline to 53.13\% (-1.78 points). In contrast, removing symbolic scoring caused a substantially larger drop to 46.25\% (-8.66 points), and the pure LLM decision-only formulation further decreased to 45.57\% (-9.34 points). These observations identify provenance-aware symbolic scoring as the dominant contributor to robust process reasoning under distribution shift.

Task-family analysis further reinforced this conclusion. In the full symbolic-stack system, the route-operation family achieved 61.19\%, the attribute-inference family reached 46.47\% and the causal-ordering family attained 96.95\%. Removing symbolic scoring produced the largest degradation in attribute inference, which decreased from 46.47\% to 34.94\% (Fig.~\ref{fig:mechanistic_ablations}B). This degradation was most pronounced for tasks requiring alignment between target steps and structurally analogous experimental precedents.

We next evaluated retrieval-view ablations under the dual-distribution-shift-trained cross-split protocol. ProvMind-Neural, which relies exclusively on encoder-based retrieval, achieved 47.04\%, 46.48\% and 46.56\% on the dual-distribution-shift, material-type and publication-year test sets, respectively. ProvMind-Symbolic, based on provenance-aware symbolic retrieval, substantially outperformed the neural-only formulation, reaching 53.65\%, 52.25\% and 54.57\%. ProvMind-Hybrid achieved 52.48\%, 51.43\% and 53.81\%.

These results suggest that encoder-only retrieval is insufficient under severe scientific distribution shift and that provenance-aware symbolic structure remains the dominant retrieval signal within the current benchmark. Nevertheless, hybrid retrieval consistently improved over the neural-only formulation by 4.95--7.25 percentage points and remained competitive with the symbolic-only system across all clean evaluation settings (Fig.~\ref{fig:mechanistic_ablations}C).

Collectively, these ablations indicate that the principal signal in MatProcBench remains encoded in provenance-aware symbolic structure, including route continuity, precursor overlap, activity compatibility and causal consistency. Neural similarity contributes complementary information, but does not yet replace explicit process-aware reasoning under realistic scientific shift. Additional full-system ablations covering reference baselines, module removals, scoring variants, retrieval decomposition, fusion-weight sweeps and top-$k$ sensitivity are reported in Supplementary Note~4.

\section{Discussion}
The results collectively suggest that materials process reasoning constitutes a distinct inference regime that cannot be reduced to conventional scientific question answering or static retrieval augmentation \citep{david2023promise}. Across both same-split and cross-split evaluations, the strongest performance consistently emerged from provenance-grounded process memory rather than from parametric recall or neural similarity alone. This distinction became most pronounced under realistic scientific distribution shift, where temporal separation and material-class separation substantially weakened the effectiveness of purely neural retrieval strategies.

These findings have several implications for computational materials science. First, they indicate that experimentally structured provenance graphs provide more than a convenient representation of synthesis workflows. Provenance structure defines the causal and relational substrate required for reasoning about process continuity, variable propagation and intervention compatibility. Many of the benchmark tasks, including route reconstruction, process ordering and condition inference, depend directly on relationships that are difficult to recover from flat textual representations alone. The results therefore support the broader view that scientific process reasoning should be grounded in structured experimental state transitions rather than treated as generic sequence modeling \citep{kononova2019text, wang2022dataset, muraoka2019linking, tsuruta2025matprov}.

Second, the observed superiority of provenance-aware symbolic structure under distribution shift suggests that scientific reasoning may require substantially stronger structural grounding than conventional retrieval-augmented generation systems typically provide \citep{lewis2020retrieval, hu2025grag, peng2025graph}. Encoder-based retrieval remained useful throughout the benchmark, particularly when combined with symbolic retrieval, but consistently underperformed provenance-grounded symbolic comparison under the cleanest evaluation settings. This pattern suggests that local semantic similarity is insufficient for robust reasoning about experimental procedures when causal dependencies and combinatorial process constraints become dominant. More broadly, these findings align with the emerging view that scientific inference often depends on explicit structural compatibility rather than lexical or embedding-level proximity alone \citep{sun2023think, jin2024graph}.

Third, the large divergence between random partitioning and shift-aware evaluation highlights the importance of evaluation protocol design for scientific reasoning benchmarks. Random splitting substantially inflated performance on tasks requiring global route discrimination or joint condition recovery, particularly A1 route retrieval and B2 full-condition-set prediction. These tasks are precisely those most vulnerable to hidden overlap in route templates or condition configurations. The present results therefore suggest that random partitioning may substantially overestimate reasoning capability in scientific process benchmarks where structural redundancy exists across related procedures. Shift-aware evaluation is consequently necessary for measuring whether a framework genuinely generalizes beyond memorized experimental patterns.

The task hierarchy observed throughout the benchmark further clarifies the current limitations of process-memory reasoning systems. Tasks involving local causal continuity or explicit process ordering remained comparatively tractable, whereas globally constrained reasoning problems such as route discrimination and joint condition recovery remained substantially more difficult. This distinction suggests that the primary unresolved challenge is not local provenance grounding itself, but global combinatorial reasoning over multi-step experimental trajectories. Future progress may therefore require stronger mechanisms for long-range process consistency, explicit constraint propagation and hierarchical experimental planning \citep{jansen2002concept, aykol2021rational, chen2024navigating}.

Several limitations should also be acknowledged. First, although MatPROV provides structured provenance graphs, the benchmark remains restricted to synthesis procedures represented within a single provenance formalism. In addition, because MatPROV is constructed through LLM-based extraction from literature, the benchmark inevitably inherits extraction noise, schema inconsistencies and occasional factual omissions or misassignments introduced during automated provenance construction. Improving the precision and robustness of literature-to-provenance extraction therefore remains an important direction for future work. Second, the current benchmark primarily evaluates reasoning over discrete process structure rather than continuous physical dynamics. Integrating process-memory reasoning with physically grounded simulation or mechanistic modeling may therefore further improve scientific validity, particularly as synthesis research increasingly couples process records with condition optimization and mechanistic constraints \citep{wang2024optimal, chen2024navigating}. Third, our supervised adaptation baseline is intentionally limited to straightforward SFT, and stronger benchmark performance may require training strategies that are more explicitly aligned with provenance-grounded reasoning, retrieval-conditioned decision making or structured constraint learning. Exploring such specialized training methods is therefore a promising avenue for future work. Fourth, while the present study focuses on materials synthesis, the underlying formulation may generalize to broader forms of scientific workflow reasoning, including biological protocols, chemical reaction systems and automated laboratory planning \citep{coley2017computer, ucak2022retrosynthetic, duigou2026retrorules, boiko2023autonomous, tom2024self}.

Taken together, the present results establish materials process informatics as a computational reasoning problem grounded in experimentally structured provenance graphs rather than in isolated textual descriptions. More broadly, they suggest that robust scientific reasoning under realistic distribution shift may depend fundamentally on process memory, causal structure and provenance-grounded experimental comparison.

\section{Methods}
\label{sec:methods}

\subsection{MatPROV provenance graphs}

MatProcBench is constructed exclusively from MatPROV provenance records \citep{tsuruta2025matprov}, which encode materials synthesis procedures as provenance graphs in PROV-JSONLD format. As a literature-derived process resource, MatPROV is related in spirit to prior text-mined and structured synthesis corpora, while extending them toward explicitly typed provenance relations \citep{kim2017materials, kononova2019text, muraoka2019linking, wang2022dataset}. Each record is parsed into a heterogeneous directed graph

\[
G=(V_m \cup V_t \cup V_a,\; E_u \cup E_g),
\]

where $V_m$, $V_t$ and $V_a$ denote material entities, tool entities and synthesis activities, respectively. A directed Usage edge $(v,a)\in E_u$ links a material or tool entity to the activity that consumes it, whereas a directed Generation edge $(a,v)\in E_g$ links an activity to the entity it produces. This representation preserves the experimentally meaningful asymmetry between material consumption and production that is typically obscured when synthesis procedures are flattened into textual sequences \citep{kononova2019text, wang2022dataset}.

Node attributes are retained directly from the source provenance graph. Material and tool entities preserve labels together with any available physical descriptors, including form, purity, mass, concentration and size-related attributes. Activity nodes preserve operation labels together with all reported process conditions, including temperature, duration, atmosphere, pressure, heating rate and rotation-related attributes when available. These condition-bearing fields are especially important because they support the kinds of synthesis-variable analyses emphasized in prior condition-prediction and process-optimization studies \citep{huo2022machine, karpovich2023interpretable, wang2024optimal}. Rather than imposing a manually simplified schema, the benchmark preserves the provenance granularity provided by MatPROV whenever the underlying JSON-LD fields can be parsed reliably.

To support process-level reasoning, topology-aware semantic roles are derived directly from graph connectivity. A material entity with outgoing usage edges but no incoming generation edge is treated as a precursor, a node with both incoming generation and outgoing usage edges is treated as an intermediate, and a node with incoming generation edges but no outgoing usage edge is treated as a product. Activity precedence is inferred from material flow: if activity $a_i$ generates an entity later consumed by activity $a_j$, then $a_i$ is constrained to precede $a_j$. Activities are subsequently topologically sorted under these constraints, while source ordering is retained only for residual ambiguities unresolved by graph structure. The resulting representation yields an executable process view in which route continuity, intermediate states and tool usage remain explicit.

\subsection{Task construction and split design}

MatProcBench is formulated as a provenance-grounded multiple-choice benchmark. We retain only synthesis graphs containing at least one activity and one precursor entity, then instantiate seven reasoning tasks whose gold answers are deterministically recoverable from provenance evidence.

The tasks are organized into three reasoning regimes. Route and local-process continuity are evaluated through A1 route retrieval, A2 missing-step identification and A3 next-activity prediction. Step-level attribute inference is evaluated through B1 condition prediction, B2 full-condition-set prediction and C tool selection. Global causal consistency is evaluated through D process ordering.

Distractors are sampled from corpus-derived task-specific candidate pools rather than synthetic random negatives. Global pools are constructed over observed routes, activity labels, tool labels, material forms, condition values and complete condition tuples. Distractors are then sampled conditionally according to task type and, when appropriate, local activity or form-transition context. This procedure preserves procedural and chemical plausibility across candidate options.

We evaluate MatProcBench under four partition protocols designed to isolate distinct generalization regimes. The random split follows a conventional 80/10/10 partition. The publication-year split enforces temporal separation, with training examples restricted to publication year $\leq 2019$, development examples drawn from 2020 and test examples drawn from $\geq 2021$. The material-type split evaluates material-class transfer by reserving battery-related synthesis records exclusively for testing while training on non-battery materials. The strictest setting is the dual-distribution-shift split, which simultaneously enforces temporal and material-class separation: training uses non-battery records published no later than 2019, development uses non-battery records from 2020 and testing uses battery-related records published in or after 2021. Records violating this separation are excluded from the corresponding partition. Supplementary Note~1 further reports split-level material-type composition, publication-year distributions and DOI overlap statistics.

Throughout this work, train, development and test refer to benchmark partitioning rather than universal parameter optimization. Only supervised fine-tuning baselines update model parameters using the training partition. Retrieval-based methods, including RAG, GraphRAG and ProvMind, instead restrict their retrieval corpora, process memory and provenance statistics to the corresponding training split while keeping backbone language-model parameters fixed.

\subsection{Baseline methods}
We compare ProvMind against four categories of baselines designed to isolate the contribution of provenance-grounded process memory. The full evaluation settings for zero-shot prompting, few-shot prompting, RAG, GraphRAG, supervised fine-tuning and ProvMind are summarized in Supplementary Note~3.

\paragraph{Prompting-based inference.}
We evaluate zero-shot and few-shot prompting with the same backbone language model used by the corresponding non-fine-tuned methods. Few-shot prompting uses in-context exemplars sampled exclusively from the corresponding training partition.

\paragraph{Static retrieval augmentation.}
We evaluate retrieval-augmented generation (RAG) using retrieved process records appended directly to the model context without explicit provenance-aware reasoning \citep{lewis2020retrieval}.

\paragraph{Graph-aware retrieval.}

We evaluate GraphRAG-style retrieval methods that augment language-model inference using locally connected process neighbourhoods extracted from provenance graphs \citep{jin2024graph, hu2025grag, peng2025graph}.

\paragraph{Supervised adaptation.}

We evaluate supervised fine-tuning (SFT) models trained on the corresponding benchmark split and evaluated without further adaptation.

\subsection{ProvMind framework}

ProvMind formulates materials process reasoning as provenance-grounded comparison against experimentally observed synthesis precedents. Rather than directly inferring answers from the held-out process alone, the framework constructs process memory exclusively from the training partition, retrieves structurally related synthesis trajectories and aggregates provenance-aware compatibility evidence across candidate options. This formulation preserves a strict train--test boundary while enabling reasoning through experimentally grounded process analogy \citep{noh2024retrieval, prein2025retro}.

\paragraph{Process memory construction.}

Each process record from the training partition is compiled into a structured process-memory representation containing topologically ordered activities, precursor and product entities, tool usages, process conditions and step-local material transitions.

From this representation, ProvMind derives several reusable statistical views of the training distribution. First, route-level activity-transition statistics are collected to capture common local process continuities across synthesis trajectories. Second, prefix-to-next-operation mappings are constructed to support continuation-style reasoning over partially observed routes. Third, all activity instances are stored within a step library together with neighbouring activities, associated tools, process conditions, material inputs and outputs, material forms and absolute route positions. These representations constitute provenance-grounded process priors derived entirely from experimentally observed synthesis records.

\paragraph{Process-memory retrieval.}

Given a query process, ProvMind retrieves a small set of structurally analogous synthesis trajectories from the training partition. Retrieval is designed to identify experimentally compatible precedents rather than merely semantically similar documents.

The symbolic retrieval view evaluates candidate processes according to activity overlap, route-length agreement and precursor-level correspondence between the query and retrieved trajectories. To improve robustness under scientific distribution shift, ProvMind additionally incorporates neural retrieval views derived from process-text embeddings and graph-structured provenance representations. Process-text embeddings are computed from linearized synthesis descriptions, whereas structure-aware embeddings are obtained from a frozen graph attention network applied to provenance graphs with text-derived node features.

The final retrieval score is computed as

\[
s_{\mathrm{ret}}(q,p)
=
\alpha s_{\mathrm{text}}(q,p)
+
\beta s_{\mathrm{struct}}(q,p)
+
\gamma s_{\mathrm{heur}}(q,p),
\]

where $s_{\mathrm{text}}$, $s_{\mathrm{struct}}$ and $s_{\mathrm{heur}}$ denote text-based, graph-structured and provenance-aware symbolic retrieval scores, respectively, and $\alpha+\beta+\gamma=1$.

This formulation allows retrieval to combine semantic similarity, propagated graph structure and explicit process compatibility within a unified process-memory framework.

\paragraph{Provenance-aware symbolic scoring.}

Retrieved synthesis precedents are subsequently converted into explicit compatibility estimates over candidate answer options. The symbolic scoring procedure is task aware and evaluates whether a candidate option remains structurally consistent with experimentally observed provenance patterns.

For route retrieval and process-ordering tasks, compatibility is estimated using activity-transition statistics and sequence-level agreement with retrieved trajectories. For missing-step and next-activity prediction tasks, scoring incorporates neighbouring activities, route position and prefix-continuation statistics. For condition and tool prediction tasks, the framework matches the target-step context against structurally compatible entries within the step library and aggregates evidence across retrieved precedents.

The resulting score therefore represents a provenance-grounded compatibility estimate between each candidate option and experimentally observed process memory rather than a generic confidence value.

ProvMind further supports neural and hybrid scoring variants. Neural scoring replaces explicit symbolic matching with embedding-based nearest-neighbour retrieval over process and step representations. Hybrid scoring combines normalized symbolic and neural compatibility estimates according to

\[
s_{\ell}
=
\lambda \hat{s}_{\ell}^{\mathrm{sym}}
+
(1-\lambda)\hat{s}_{\ell}^{\mathrm{neu}},
\]

where $s_{\ell}$ denotes the final compatibility score for candidate option $\ell$.

\paragraph{Language-model-mediated inference.}

Following retrieval and symbolic scoring, ProvMind provides the backbone language model with the query question, retrieved process summaries and provenance-grounded compatibility evidence. Optionally, the framework may also generate a short intermediate reasoning plan describing how retrieved precedents should be used during inference. Unlike unrestricted chain-of-thought prompting, this planning stage remains intentionally lightweight and tightly coupled to the retrieved provenance evidence.


\section*{Data availability}
The benchmarks introduced in this study are available at \url{https://github.com/ZHymLumine/MatProcBench}. The source corpora from which these resources were derived are publicly available from their original providers: MatPROV at \url{https://huggingface.co/datasets/MatPROV-project/MatPROV}. Any use of the source datasets remains subject to the terms and access conditions specified by their respective providers.

\section*{Code availability}
Code for ProvMind is available at \url{https://github.com/ZHymLumine/MatProcBench}.

\section*{Acknowledgements}
This work is supported by JST-CREST (JPMJCR21O2).

\section*{Author contributions}

Y.Z. conceived the study, constructed the benchmark, implemented the methods, performed the experiments and drafted the manuscript. K.T. supervised the project, contributed to the study design and revised the manuscript.

\section*{Competing interests}

The authors declare no competing interests.

\clearpage
\appendix
\renewcommand{\thefigure}{S\arabic{figure}}
\renewcommand{\thetable}{S\arabic{table}}
\setcounter{figure}{0}
\setcounter{table}{0}
\captionsetup{font=small,labelfont=bf}
\setlength{\parskip}{0.5em}
\setlength{\parindent}{0pt}
\let\table\tableorg
\let\endtable\endtableorg
\let\sidewaystable\sidewaystableorg
\let\endsidewaystable\endsidewaystableorg

\section*{Supplementary Information}
We provide this supplementary document to extend the quantitative and methodological details presented in the main manuscript.

\section*{Supplementary Note 1. Benchmark composition and split validity}

We summarize the benchmark composition, the material-type separation induced by the type-based and dual-OOD splits, the publication-year boundaries used for temporal evaluation, and the DOI overlap structure across train--test pairings. MatProcBench contains 34,975 multiple-choice instances spanning seven reasoning tasks.

\begin{table}[htbp]
\centering
\caption{Overall task composition of MatProcBench. Shares are computed over all 34,975 benchmark instances.}
\label{tab:supp_task_counts}
\begin{tabular}{lrr}
\toprule
Task & Count & Share (\%) \\
\midrule
A1 Route Retrieval & 1938 & 5.54 \\
A2 Missing Step & 6231 & 17.82 \\
A3 Next Activity & 7299 & 20.87 \\
B1 Condition Prediction & 12339 & 35.28 \\
B2 Full Condition Set & 1087 & 3.11 \\
C1 Tool Selection & 4390 & 12.55 \\
D Process Ordering & 1691 & 4.83 \\
\midrule
Total & 34975 & 100.00 \\
\bottomrule
\end{tabular}
\end{table}

\begin{table}[p]
\centering
\caption{Split-level overview computed from the processed benchmark files. Material percentages are reported over question instances in each partition.}
\label{tab:supp_split_overview}
\small
\setlength{\tabcolsep}{4pt}
\renewcommand{\arraystretch}{1.05}
\begin{tabular}{llrrrrrr}
\toprule
Split & Partition & Samples & Unique DOIs & Battery (\%) & Thermoelectric (\%) & Magnetic (\%) & Year range \\
\midrule
Year+Type & Train & 23654 & 1058 & 0.00 & 53.26 & 41.01 & 1982--2019 \\
Year+Type & Dev & 2970 & 129 & 0.00 & 61.18 & 32.49 & 2020 \\
Year+Type & Test & 2479 & 86 & 100.00 & 0.00 & 0.00 & 2021--2024 \\
Year+Type & Excluded & 5872 & 214 & 26.23 & 48.77 & 16.43 & 2016--2024 \\
Random & Train & 27980 & 1480 & 11.52 & 49.43 & 33.17 & 1982--2024 \\
Random & Dev & 3497 & 1209 & 11.32 & 49.16 & 33.06 & 1982--2024 \\
Random & Test & 3498 & 1191 & 11.46 & 49.46 & 34.08 & 1982--2024 \\
Material Type & Train & 27861 & 1354 & 0.00 & 55.90 & 37.43 & 1982--2024 \\
Material Type & Dev & 3095 & 1088 & 0.00 & 55.12 & 38.80 & 1982--2024 \\
Material Type & Test & 4019 & 133 & 100.00 & 0.00 & 0.00 & 2016--2024 \\
Publication Year & Train & 24838 & 1094 & 4.77 & 50.72 & 39.05 & 1982--2019 \\
Publication Year & Dev & 3326 & 140 & 10.70 & 54.63 & 29.01 & 2020 \\
Publication Year & Test & 6811 & 253 & 36.40 & 42.05 & 14.17 & 2021--2024 \\
\bottomrule
\end{tabular}
\end{table}

\begin{figure}[p]
\centering
\includegraphics[width=\textwidth]{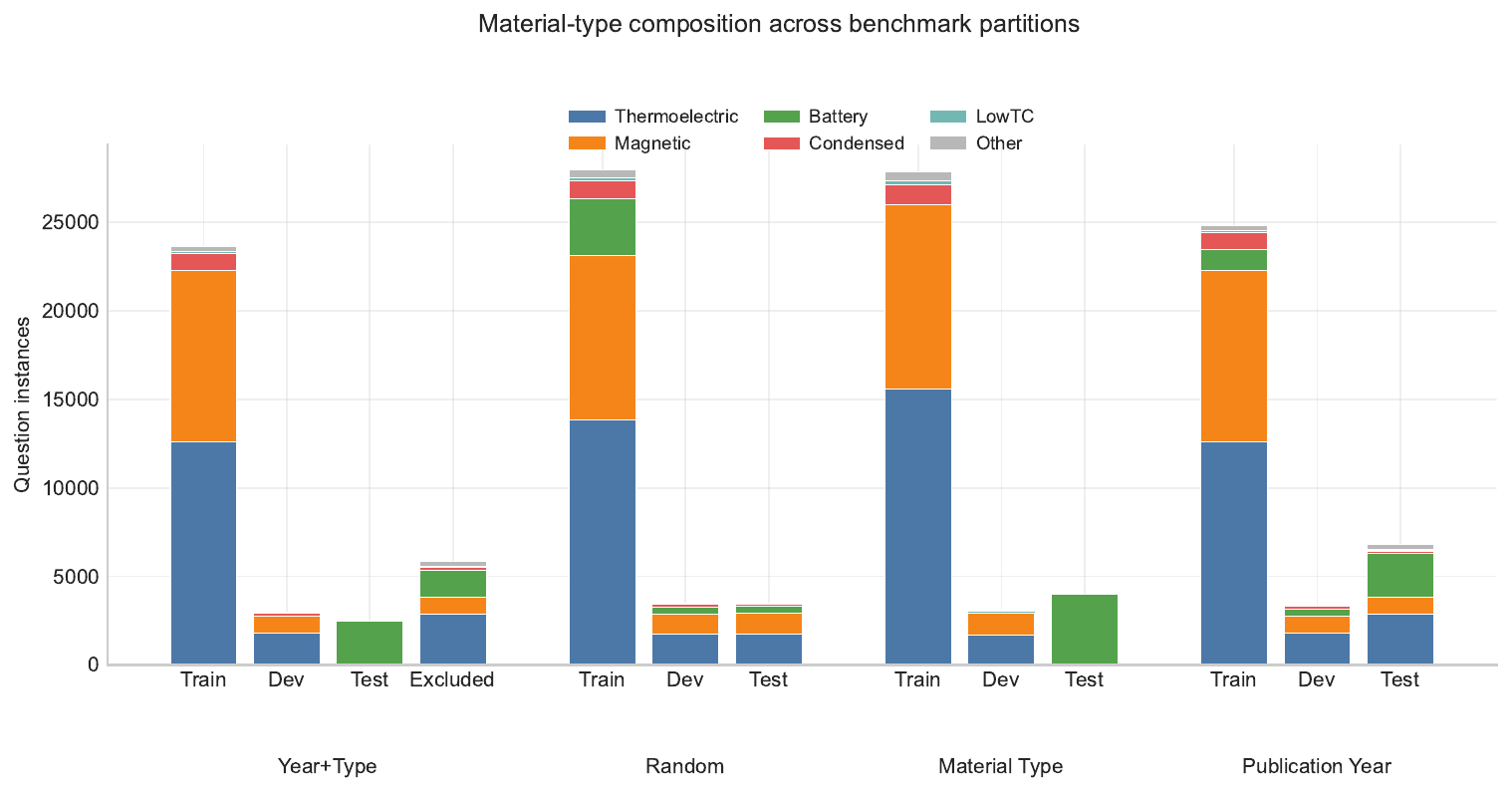}
\caption{Material-type composition across benchmark partitions. The type-based split isolates Battery records in the test partition, while the dual-OOD split further combines this material separation with a publication-year boundary.}
\label{fig:supp_material_type}
\end{figure}

\begin{figure}[p]
\centering
\includegraphics[width=\textwidth]{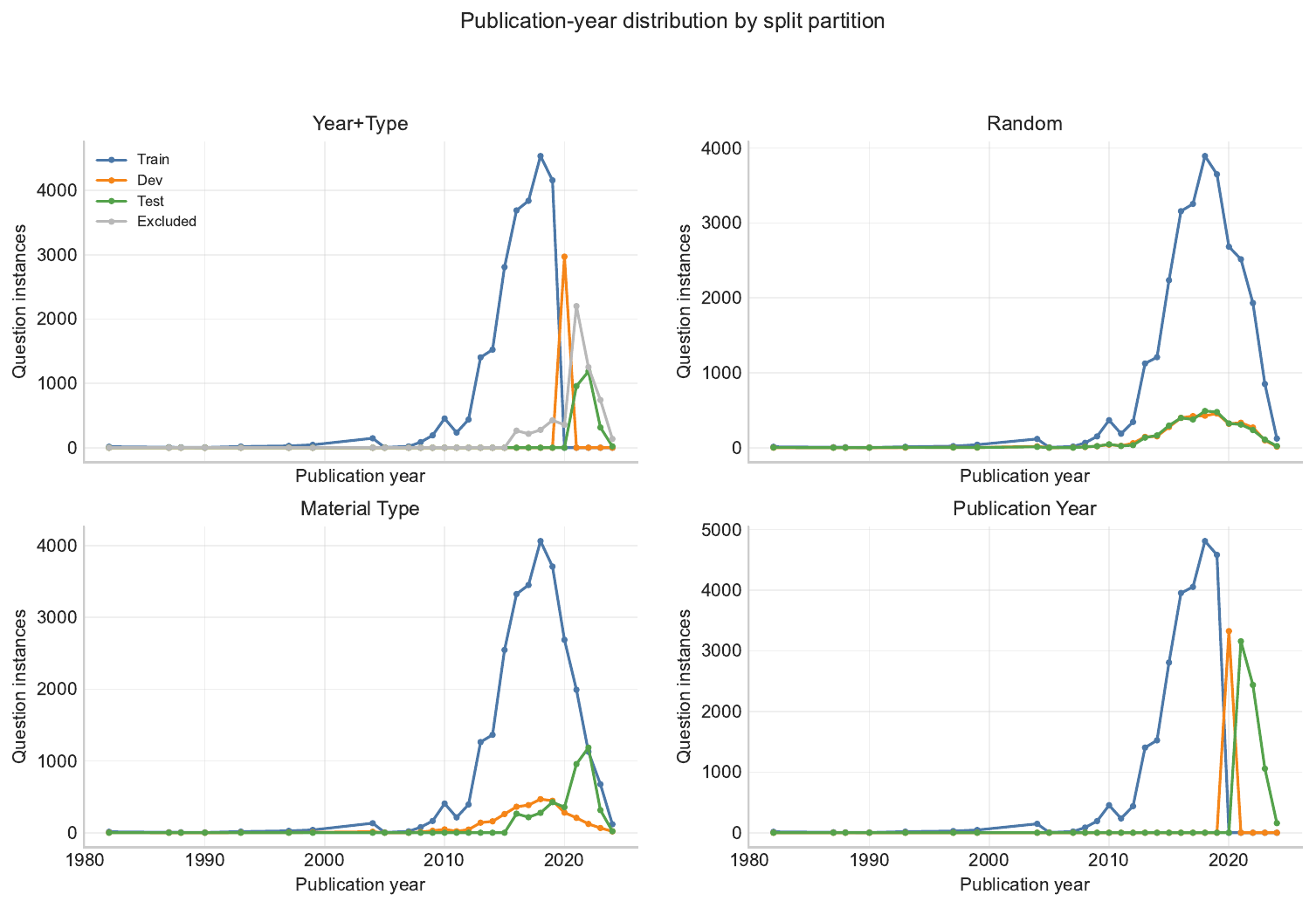}
\caption{Publication-year distribution of question instances in each split partition. The year-based split enforces train $\leq 2019$, dev $= 2020$, and test $\geq 2021$, while the dual-OOD split applies the same temporal rule only within the non-Battery training domain and the Battery test domain.}
\label{fig:supp_year_distribution}
\end{figure}

\begin{figure}[p]
\centering
\includegraphics[width=0.82\textwidth]{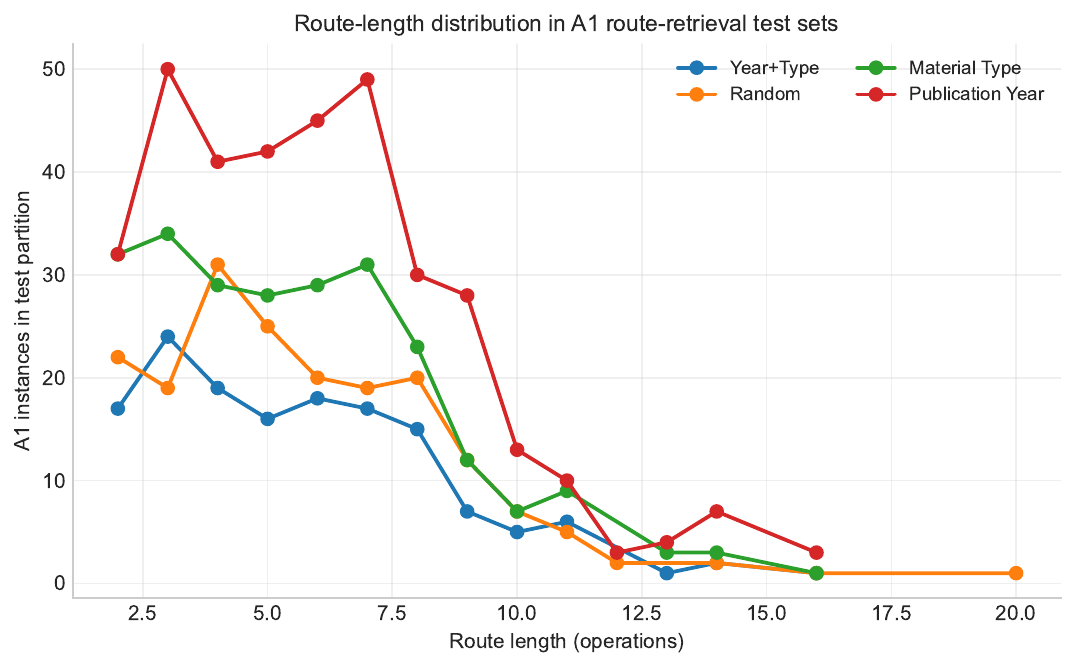}
\caption{Route-length distribution for the A1 route-retrieval test instances in each evaluation split. Route length is measured as the number of recorded operations in the gold synthesis trajectory.}
\label{fig:supp_route_length}
\end{figure}

\begin{figure}[p]
\centering
\includegraphics[width=0.72\textwidth]{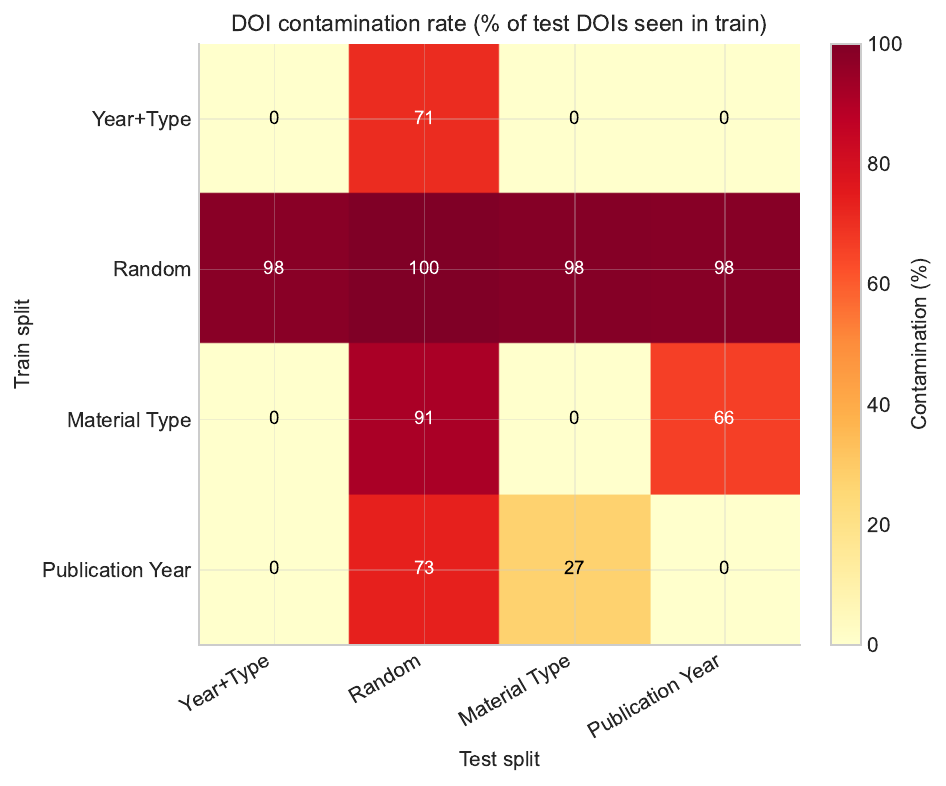}
\caption{DOI contamination matrix computed from the processed split files. Each entry reports the fraction of test DOIs that are also present in the corresponding train partition. The clean OOD pairs emphasized in the manuscript appear as the zero-contamination cells linking the dual-OOD train split to the dual-OOD, type-based, and year-based test sets.}
\label{fig:supp_contamination}
\end{figure}

\clearpage
\section*{Supplementary Note 2. Extended benchmark results}

We present the full result tables supporting the main-text performance summaries. The same-split table compares the principal prompting, retrieval, and ProvMind settings across all four benchmark partitions. The cross-split table summarizes the clean OOD setting in which train-dependent components are fixed on the dual-OOD training split and evaluated on the three non-random test partitions.

\begin{table}[htbp]
\centering
\caption{Overall same-split accuracy (\%) for the methods summarized in the main text.}
\label{tab:supp_same_split}
\small
\setlength{\tabcolsep}{6pt}
\begin{tabular}{lrrrrr}
\toprule
Split & Zero-shot & Few-shot & RAG & GraphRAG & ProvMind-Hybrid \\
\midrule
Year+Type & 45.74 & 49.33 & 43.32 & 45.70 & 52.84 \\
Random & 44.85 & 49.29 & 59.12 & 46.17 & 85.33 \\
Material Type & 44.74 & 49.42 & 42.27 & 44.74 & 52.55 \\
Publication Year & 45.47 & 48.95 & 48.50 & 45.24 & 54.82 \\
\bottomrule
\end{tabular}
\end{table}

\begin{table}[htbp]
\centering
\caption{Clean OOD cross-split accuracy (\%) when train-dependent components are fixed on the dual-OOD training split.}
\label{tab:supp_cross_split}
\small
\setlength{\tabcolsep}{4.5pt}
\begin{tabular}{lrrrrrr}
\toprule
Test split & Few-shot & RAG & GraphRAG & ProvMind-Neural & ProvMind-Symbolic & ProvMind-Hybrid \\
\midrule
Year+Type & 49.33 & 43.32 & 45.70 & 47.04 & 53.65 & 52.48 \\
Material Type & 48.84 & 42.27 & 44.94 & 46.48 & 52.25 & 51.43 \\
Publication Year & 49.05 & 43.86 & 45.35 & 46.56 & 54.57 & 53.81 \\
\bottomrule
\end{tabular}
\end{table}

\begin{table}[p]
\centering
\caption{Per-task same-split accuracy (\%) for ProvMind-Hybrid across all four benchmark partitions.}
\label{tab:supp_taskwise_hybrid}
\small
\setlength{\tabcolsep}{5pt}
\begin{tabular}{lrrrr}
\toprule
Task & Year+Type & Random & Material Type & Publication Year \\
\midrule
A1 Route Retrieval & 35.14 & 80.11 & 30.29 & 33.89 \\
A2 Missing Step & 64.81 & 79.10 & 64.79 & 70.27 \\
A3 Next Activity & 64.15 & 92.22 & 63.81 & 67.79 \\
B1 Condition Prediction & 46.86 & 88.63 & 46.79 & 46.07 \\
B2 Full Condition Set & 25.24 & 85.25 & 20.36 & 23.58 \\
C1 Tool Selection & 43.61 & 79.82 & 45.60 & 48.28 \\
D Process Ordering & 62.60 & 74.23 & 66.51 & 66.36 \\
\bottomrule
\end{tabular}
\end{table}

\begin{figure}[p]
\centering
\includegraphics[width=0.82\textwidth]{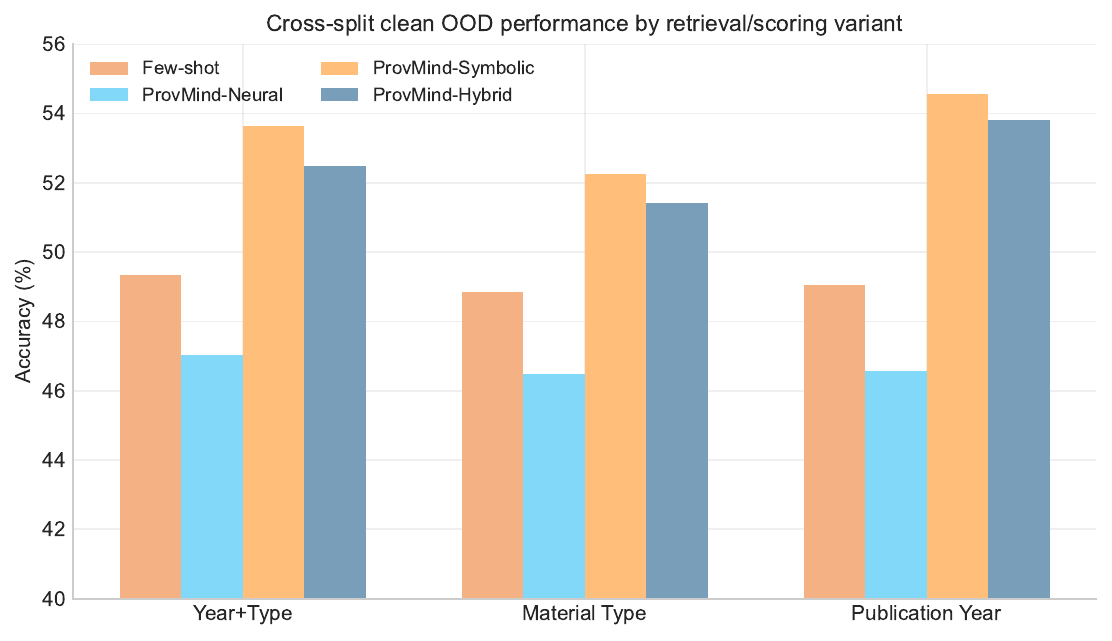}
\caption{Clean OOD cross-split comparison of few-shot prompting and the three ProvMind retrieval/scoring views. Provenance-aware symbolic structure remains the strongest single view, while the hybrid variant remains competitive across all three evaluation settings.}
\label{fig:supp_cross_split_variants}
\end{figure}

\clearpage
\section*{Supplementary Note 3. Experimental settings}

We summarize the principal experimental settings for zero-shot prompting, few-shot prompting, RAG, GraphRAG, supervised fine-tuning, and ProvMind.

\begin{table}[p]
\centering
\caption{Experimental settings for the baseline and ProvMind evaluations reported in this study.}
\label{tab:supp_hyperparameters}
\small
\setlength{\tabcolsep}{4pt}
\renewcommand{\arraystretch}{1.08}
\centering
\begin{subtable}[t]{0.98\textwidth}
\centering
\caption*{\textbf{Shared}}
\begin{tabular}{p{0.34\linewidth}p{0.60\linewidth}}
\toprule
Setting & Value \\
\midrule
Backbone model & Qwen/Qwen2.5-7B-Instruct \\
Inference precision & bfloat16 with automatic device mapping \\
\bottomrule
\end{tabular}
\end{subtable}

\vspace{0.75em}

\begin{subtable}[t]{0.48\textwidth}
\centering
\caption*{\textbf{Zero-shot}}
\begin{tabular}{p{0.34\linewidth}p{0.60\linewidth}}
\toprule
Setting & Value \\
\midrule
Prompting setup & Question-only prompting with greedy decoding \\
Generation budget & 16 new tokens \\
\bottomrule
\end{tabular}
\end{subtable} \hfill \begin{subtable}[t]{0.48\textwidth}
\centering
\caption*{\textbf{Few-shot}}
\begin{tabular}{p{0.34\linewidth}p{0.60\linewidth}}
\toprule
Setting & Value \\
\midrule
In-context exemplars & 3 task-matched training examples \\
Sampling seed & 42 \\
Generation budget & 16 new tokens \\
\bottomrule
\end{tabular}
\end{subtable}

\vspace{0.75em}

\begin{subtable}[t]{0.48\textwidth}
\centering
\caption*{\textbf{RAG}}
\begin{tabular}{p{0.34\linewidth}p{0.60\linewidth}}
\toprule
Setting & Value \\
\midrule
Retriever & FAISS dense retrieval over all-mpnet-base-v2 embeddings \\
Retrieved items & Top-3 training records \\
Generation budget & 16 new tokens \\
\bottomrule
\end{tabular}
\end{subtable} \hfill \begin{subtable}[t]{0.48\textwidth}
\centering
\caption*{\textbf{GraphRAG}}
\begin{tabular}{p{0.34\linewidth}p{0.60\linewidth}}
\toprule
Setting & Value \\
\midrule
Retriever & Graph-aware retrieval with graph\_k = 3 and graph\_hops = 1 \\
Text encoder & all-mpnet-base-v2 \\
Generation budget & 16 new tokens \\
\bottomrule
\end{tabular}
\end{subtable}

\vspace{0.75em}

\begin{subtable}[t]{0.48\textwidth}
\centering
\caption*{\textbf{SFT (same-split)}}
\begin{tabular}{p{0.34\linewidth}p{0.60\linewidth}}
\toprule
Setting & Value \\
\midrule
Training strategy & Evidence-augmented LoRA fine-tuning \\
Optimization & batch size = 8, gradient accumulation = 4, learning rate = 2e-4, warmup = 50 \\
LoRA setup & r = 16, alpha = 32, max sequence length = 1024 \\
Task reweighting & B1\_condition\_prediction x2; D1\_process\_ordering x2 \\
\bottomrule
\end{tabular}
\end{subtable} \hfill \begin{subtable}[t]{0.48\textwidth}
\centering
\caption*{\textbf{SFT (cross-split)}}
\begin{tabular}{p{0.34\linewidth}p{0.60\linewidth}}
\toprule
Setting & Value \\
\midrule
Training strategy & LoRA fine-tuning for clean OOD transfer \\
Optimization & max steps = 3000, batch size = 4, gradient accumulation = 4, learning rate = 2e-4, warmup = 200 \\
LoRA setup & r = 32, alpha = 64, max sequence length = 1024 \\
Data balancing & balance strategy = sqrt; evidence dropout = 0.0 \\
\bottomrule
\end{tabular}
\end{subtable}

\vspace{0.75em}

\begin{subtable}[t]{0.98\textwidth}
\centering
\caption*{\textbf{ProvMind}}
\begin{tabular}{p{0.34\linewidth}p{0.60\linewidth}}
\toprule
Setting & Value \\
\midrule
Text encoder & all-mpnet-base-v2 \\
Retrieval depth & Top-8 retrieved training processes \\
Retrieval fusion & $\alpha = 0.4$, $\beta = 0.3$, $\gamma = 0.3$ \\
Same-split scoring & Symbolic compatibility scoring \\
Cross-split scoring & Hybrid scoring \\
Hybrid score weights & symbolic = 0.5, neural = 0.5 \\
Generation budgets & 96 new tokens for planning and 48 new tokens for final answer generation \\
\bottomrule
\end{tabular}
\end{subtable}

\vspace{0.75em}

\begin{subtable}[t]{0.98\textwidth}
\centering
\caption*{\textbf{Split design}}
\begin{tabular}{p{0.34\linewidth}p{0.60\linewidth}}
\toprule
Setting & Value \\
\midrule
Type split & Battery records reserved for test \\
Year split & train $\leq$ 2019, dev = 2020, test $\geq$ 2021 \\
Dual-OOD split & train non-Battery $\leq$ 2019; dev non-Battery = 2020; test Battery $\geq$ 2021 \\
\bottomrule
\end{tabular}
\end{subtable}
\end{table}

\clearpage
\section*{Supplementary Note 4. Extended ablations of ProvMind}

We report additional full-system ablations of ProvMind on the dual-distribution-shift same-split setting (Year+Type) using Qwen2.5-7B-Instruct. Whereas the main text focuses on symbolic-stack ablations within ProvMind-Symbolic and clean-OOD retrieval-view comparison, the tables below summarize complementary design-space analyses for the complete ProvMind framework, including reference baselines, module removals, scoring variants, retrieval decomposition, fusion-weight sweeps and top-$k$ sensitivity.

\begin{table}[p]
\centering
\caption{Extended full-system ablations of ProvMind on the dual-distribution-shift same-split setting (Year+Type) with Qwen2.5-7B-Instruct. The reference block compares the pure language-model baseline, the symbolic-only ProvMind variant, and the full hybrid system. The module and scoring blocks summarize full-system design choices beyond the symbolic-stack ablations shown in the main text.}
\label{tab:supp_extended_ablation_core}
\small
\setlength{\tabcolsep}{4.5pt}
\renewcommand{\arraystretch}{1.08}
\begin{tabular}{llrr}
\toprule
Group & Configuration & Accuracy (\%) & Correct / Total \\
\midrule
\multirow{3}{*}{Reference} 
& LLM-only & 46.39 & 1150 / 2479 \\
& ProvMind-Symbolic & 53.65 & 1330 / 2479 \\
& ProvMind-Hybrid & 52.68 & 1306 / 2479 \\
\midrule
\multirow{4}{*}{Modules}
& ProvMind-Hybrid w/o planning & 51.07 & 1266 / 2479 \\
& ProvMind-Hybrid w/o symbolic fallback & 52.68 & 1306 / 2479 \\
& ProvMind-Hybrid w/o symbolic scoring & 47.04 & 1166 / 2479 \\
& ProvMind-Hybrid w/o planning and symbolic fallback & 51.07 & 1266 / 2479 \\
\midrule
\multirow{5}{*}{Scoring}
& ProvMind-Neural scoring & 38.85 & 963 / 2479 \\
& ProvMind-Symbolic scoring & 53.09 & 1316 / 2479 \\
& ProvMind-Hybrid scoring (0.5 sym / 0.5 neu) & 52.68 & 1306 / 2479 \\
& ProvMind-Hybrid scoring (0.7 sym / 0.3 neu) & 52.88 & 1311 / 2479 \\
& ProvMind-Hybrid scoring (0.3 sym / 0.7 neu) & 53.01 & 1314 / 2479 \\
\bottomrule
\end{tabular}
\end{table}

\begin{table}[p]
\centering
\caption{Retrieval and sensitivity ablations for ProvMind on the dual-distribution-shift same-split setting (Year+Type) with Qwen2.5-7B-Instruct. These results summarize retrieval-view decomposition, fusion-weight sweeps and top-$k$ sensitivity within the full ProvMind design space.}
\label{tab:supp_extended_ablation_sensitivity}
\small
\setlength{\tabcolsep}{4.5pt}
\renewcommand{\arraystretch}{1.08}
\begin{tabular}{llrr}
\toprule
Group & Configuration & Accuracy (\%) & Correct / Total \\
\midrule
\multirow{7}{*}{Retrieval}
& Text only & 53.41 & 1324 / 2479 \\
& Structure only & 52.04 & 1290 / 2479 \\
& Heuristic only & 52.76 & 1308 / 2479 \\
& Text + structure & 52.12 & 1292 / 2479 \\
& Text + heuristic & 52.44 & 1300 / 2479 \\
& Structure + heuristic & 52.44 & 1300 / 2479 \\
& Full default retrieval & 52.68 & 1306 / 2479 \\
\midrule
\multirow{4}{*}{Fusion}
& Equal weights & 52.00 & 1289 / 2479 \\
& Text-heavy fusion & 52.44 & 1300 / 2479 \\
& Structure-heavy fusion & 52.56 & 1303 / 2479 \\
& Heuristic-heavy fusion & 52.00 & 1289 / 2479 \\
\midrule
\multirow{5}{*}{Top-$k$}
& $k = 1$ & 52.96 & 1313 / 2479 \\
& $k = 2$ & 52.76 & 1308 / 2479 \\
& $k = 4$ & 52.80 & 1309 / 2479 \\
& $k = 8$ & 52.68 & 1306 / 2479 \\
& $k = 16$ & 52.60 & 1304 / 2479 \\
\bottomrule
\end{tabular}
\end{table}

\end{document}